\documentclass{IOS-Book-Article}

\usepackage{mathptmx}
\usepackage{tikz}
\usepackage{subcaption}

\usepackage{algorithm2e}
 \usepackage{algpseudocode}
 \usepackage{listings}
\newtheorem{defn}{Definition}
\usepackage{soul, color}\setuldepth{article}
  \usepackage{url} 
  \usepackage{graphicx}
\def\hb{\hbox to 10.7 cm{}}

\begin{document}

\pagestyle{headings}
\def\thepage{}

\begin{frontmatter}              % The preamble begins here.

%\pretitle{Pretitle}
% Fuzzy Sets and Color Harmony: An Examination of Universality
% Exploring Color Harmony: A Fuzzy Sets Approach and Universality Assessment
% Color Harmony Evaluation through Fuzzy Logic: An Investigation of Universality
% Fuzzy Sets in Color Harmony Evaluation: Uncovering Universal Principles
% Fuzzy Logic Approach Towards Color Harmony
% The Universality of Color Harmony: A Fuzzy Sets Perspective
% Evaluating Color Harmony Universality Using Fuzzy Sets

\title{Towards a Universal Understanding of Color Harmony: Fuzzy  Approach}

%On color hamrmony universality
\markboth{}{September 2023\hb}
%\subtitle{Subtitle}

\author[A]{\fnms{Pakizar} \snm{Shamoi}%
\thanks{Corresponding Author: Pakizar Shamoi; E-mail:
p.shamoi@kbtu.kz}},
\author[A]{\fnms{Muragul } \snm{Muratbekova}},
\author[A]{\fnms{Assylzhan} \snm{Izbassar}},
\author[B]{\fnms{Atsushi} \snm{Inoue}}
and
\author[C]{\fnms{Hiroharu} \snm{Kawanaka}}

%\address[a]{}
% \author[a]{ } 
% \author[a]{}

%\runningauthor{B.P. Manager et al.}
\address[A]{School of Information Technology and Engineering, Kazakh-British Technical University, Almaty, Kazakhstan}
\address[B]{Kyushu Institute of Technology, Kitakyushu, Japan}
\address[C]{Graduate School of Engineering, Mie University, Tsu, Japan}

% Harmony level prediction is receiving more and more attention nowadays. Color is a crucial factor affecting human aesthetic response. Our perception is highly subjective and context-dependent, however, some sort of universality can also take place in this field of study. In this paper, we narrow into several different contexts  to trace certain patterns in each of them. Exploring these correlations using fuzzy logic can provide more relationships between color palettes, context, and image aesthetic level. 
\begin{abstract}
% Nowadays, harmony level prediction is receiving increasing attention.%
Harmony level prediction is receiving increasing attention nowadays. Color plays a crucial role in affecting human aesthetic responses. In this paper, we explore color harmony using a fuzzy-based color model and address the question of its universality. For our experiments, we utilize a dataset containing attractive images from five different domains: fashion, art, nature, interior design, and brand logos. We aim to identify harmony patterns and dominant color palettes within these images using fuzzy approach. It is well-suited for this task because it can handle the inherent subjectivity and contextual variability associated with aesthetics and color harmony evaluation. Our experimental results suggest that color harmony is largely universal. Additionally, our findings reveal that color harmony is not solely influenced by hue relationships on the color wheel but also by the saturation and intensity of colors. In palettes with high harmony levels, we observed a prevalent adherence to color wheel principles while maintaining moderate levels of saturation and intensity. These findings contribute to ongoing research on color harmony and its underlying principles, offering valuable insights for designers, artists, and researchers in the field of aesthetics.

\end{abstract}

\end{frontmatter}
%\markboth{January 2020\hb}{January 2020\hb}
%\thispagestyle{empty}
%\pagestyle{empty}

\section{Introduction}

%Despite the fact that we are surrounded by aesthetic experiences, no scientifically comprehensive theory can explain, evaluate, or predict aesthetic preferences. 
 
 % For many years, researchers have been exploring image features to measure the aesthetic qualities of objects. No such universal features have been discovered so far, . 
 %There is limited research on the universality of color harmonies. 
 
% \hl{Introduce the subject with hook. CH is important}
The human brain tends to seek a visually harmonious experience. With an increase of digital affective information more researchers are interested in objective aesthetic level assessment %\cite{cnn}. 
Automatic prediction of image harmony value is receiving more attention \cite{2022_1}, however, still, there is no well-constructed theory to use as a guidance for practical purposes \cite{gregor1},\cite{fss}, \cite{scis2023}.
% \hl{Jistify. No tools to evaluate CH.}
One reason for this is that aesthetic levels can vary across different domains \cite{domain}. Moreover, human perception is subjective by nature \cite{lit2}. %so for this article, we rely on the color dimension as it is the most influential visual factor.

% \hl{state research questions or hypothesis.color harmony and its rules across contexts; color harmony metrics that take into account intensity and saturation, not only hue;}
Color harmony is the primary driver of aesthetic preference for color schemes \cite{palmer_ar}. Several researches have shown that color harmonies can be universal \cite{lit1}. Such combinations as monochromatic, complementary, analogous, etc. are widely used in art, fashion, and interior design. However, most of them only use hue as a parameter, whereas color should be described by several parameters (for example, hue, saturation, and intensity). On the other hand, some studies \cite{jacob}, \cite{Amsteus2015a} show that the aesthetics level is highly context-specific and changes across different domains. So, further research is required in order to resolve these contradictions. 

% identify whether color harmony is universal or situational?
% \hl{The contributions of this [paper are:]}
This paper aims to comprehend the extent of color harmony universality. Our study goes beyond traditional approaches by considering the role of saturation and intensity alongside hue in color harmony assessment. In our earlier studies, we introduced the fuzzy color model (FHSI) \cite{fss,ijcai,ijufks,jaciii} that can be used to address visual uncertainty. This study employs FHSI to investigate color harmony universality. Whether color palettes that are considered harmonious in one context are likewise considered harmonious in other contexts? Most studies focus on harmonies within 2 or 3-color palettes, but in real-world scenarios, we often deal with much more complex palettes. In our research, we analyze 8-color palettes in five domains (nature, fashion, art, logo and interior design) to better reflect the complexity found in practical applications.

% [Assylzhan] In the above sentence where used 8-color palette, maybe here should be 9-color palette?

%The contributions of our paper are a set of rules on color aesthetics that can be widely used as practical guidance and further research.

% [Assylzhan] How you look like to "We can consider this paper structure as, firstly introduction, where we describe color harmony limitation, following with provided overviews of previous studies on color harmony. Next, we recall basic ideas from our previous research works on fuzzy color space from Section I to III, consequently. Starting with Section IV, we described the methods and approchaes to made an experiments, as well as a dataset description. The subsequent Section V presents experimental result, and finally, the last section concludes the paper, and provides some future work ideas to modify our constructed methodology."

The paper is structured as follows. Section I is this introduction. Section II provides an overview of previous studies on color harmony. Next, we recall basic ideas from our previous works on fuzzy color space in Section III. Section IV describes the methods we use in this study, as well as the dataset description. The subsequent section, Section V, presents experimental results. Finally, Section VI concludes the paper and provides recommendations for future methodology improvement.

%\subsection{Problem Statement}
%1. Introduce specific subject of research. Premises. Grab the attention. definitions. meaning, importance
%Justyfy the choice of the topc. The info sentences that connect grabber to thesis statements - key points, back info
%Previous research revealed... significance
%3.state the objectives of the inversigation, research questions or hypothesis
%the thesis part - states the main idea of the paper

\section{Related Work}
% \hl{Literature review is here, with critique. Logic of grouping: by time?}
The harmony (pleasantness) of the color scheme is very important in color aesthetics \cite{Ou2009,fss, scis2023}. The study of color harmony has a long history. None of the methods, however, were found to be appropriate \cite{Ou2009}. The most common method of creating harmony is likely a selection of colors from a color wheel as recommended by Goethe  \cite{goethe}  and Itten \cite{itten}. Alternatively, other approaches to determine color harmony, proposed by Moon and Spencer  \cite{moon} and Chevreul \cite{chevreu2}, are based on the examination of color relationships. Typically, these studies operate on the premise that colors achieve harmony when they are either complementary or analogous (similar). Other methods include Matsuda's color coordination \cite{tokumaru}, and deep learning approaches \cite{dl}.

% [Assylzhan] Here, how about some changes in the last sentence of the above paragraph?

The other important question is color harmony universality. Many researchers concluded that it is highly context-dependent \cite{colorpsycho}. This specificity varies from the application field we observe to the viewer's personal condition and subjective judgments. Color preference can also be influenced by different factors, such as gender, age, sex, and geographical region \cite{compar}. At the same time, some studies obtained in their experiments certain rules of color aesthetic universality \cite{2022}, \cite{scis2023}. Specifically, some color combinations tend to arouse similar human responses in whatever context is given.

From the studies mentioned above, we can see that the mechanisms underlying the uncertainty of color harmonies remain controversial. As aesthetic-level prediction receives more attention, it is crucial to understand the patterns and rules that form the basis of this area of interest. 
% \hl{As a result, claims have been made Consequently, assertions have been made that there were no laws governing color harmon \cite{Ou2009}. }

\section{Research Background: Fuzzy Color Modelling}
\begin{figure}[tb]
 \centering
\includegraphics[width=3.5in]{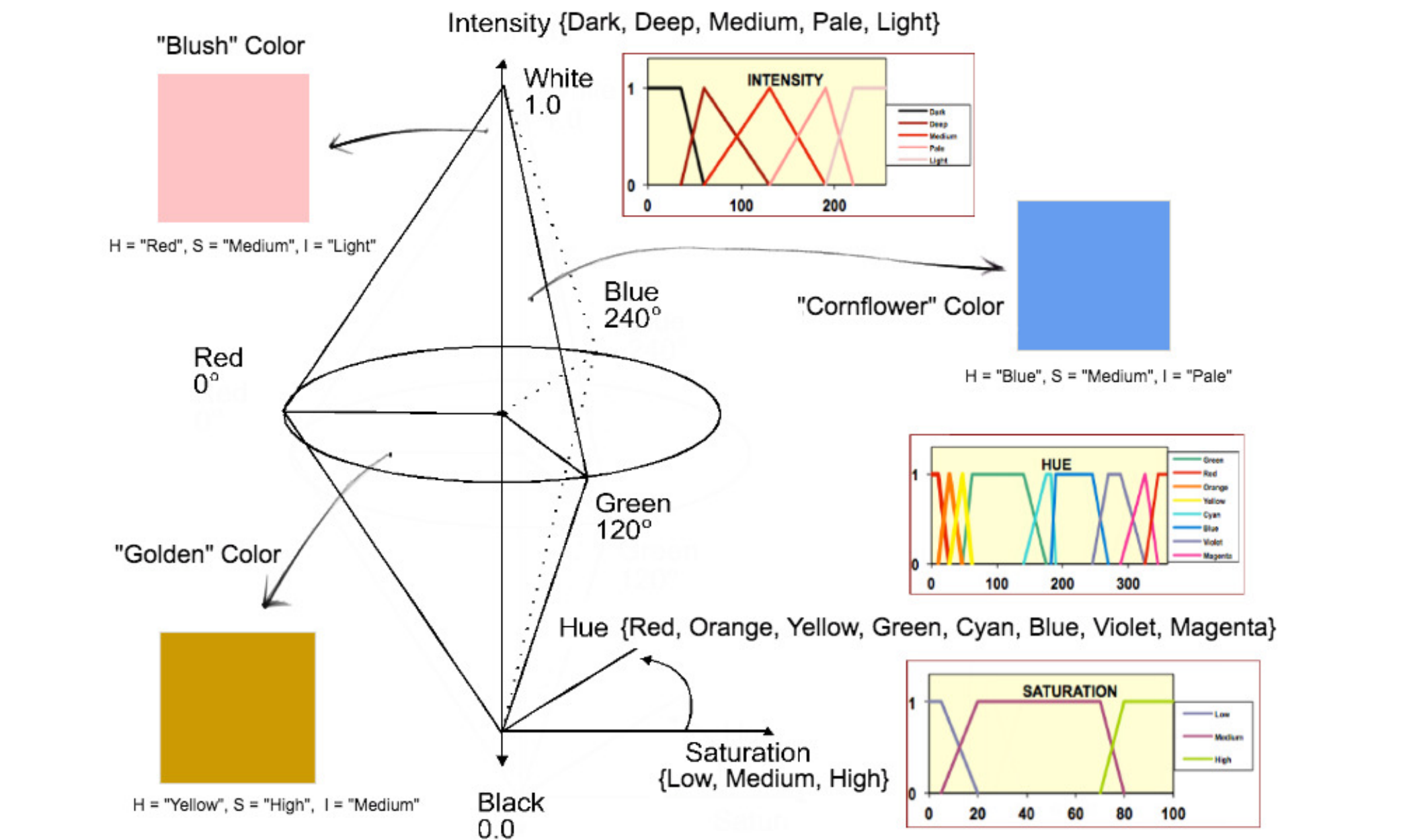}

\caption{FHSI Color Space. Hue, Saturation, and Intensity attributes are represented as fuzzy sets.}
\label{hsimodel}
\end{figure}

%\subsubsection{Fuzzy Color Modelling}
In our earlier works \cite{fss,jaciii,ijufks,ojis,ijcai},  we presented a novel fuzzy sets based color space, FHSI, which is consistent with human perception. Our method relies on fuzzifying the well-known HSI color space.  (see Table \ref{tab:fhsi} and Figure \ref{hsimodel} for more details). We also provided objective measures for finding the image similarity to match human evaluation \cite{thesis, fss}. So, fuzzy color is a fuzzy subset of points of some crisp color space\cite{soto}, which is the HSI space in our case \cite{thesis, fss}. Let $ D_{H} $, $ D_{S} $, $ D_{I} $ be domains of the $H$, $S$, $I$ attributes respectively. 

\begin{defn}
\label{def1}
 \textbf{FHSI (fuzzy HSI) color} $C$ is a linguistic label whose semantic is represented in HSI color space by a normalized fuzzy subset of $ D_{H} \times  D_{S}  \times D_{I} $.
\end{defn}
 
    From Definition~\ref{def1} it is obvious that for each fuzzy color $C$ there exists at least one representative crisp color whose membership to $C$ is 1. Now let's extend the concept of fuzzy color to a concept of a fuzzy color space. %Using the basic definition of a fuzzy color space\cite{soto}, in our case:
    
\begin{defn}
\label{def2}
 \textbf{FHSI (fuzzy HSI) color space} is a set of fuzzy colors that define a partition of $ D_{H} \times  D_{S}  \times D_{I} $.
\end{defn}

In other words, a fuzzy color space is a collection of fuzzy sets that provides a conceptual quantization (with soft boundaries) of crisp color space \cite{soto}.

\begin{defn}
\label{def3}
\textbf{ FHSI (fuzzy HSI) color palette} is a combination of several fuzzy colors. 
\end{defn}
So, in a fuzzy color palette, each element is not a crisp color (color point), but a fuzzy color (region), e.g., for the \textit{Blush} color, H = \textit{Red}, S = \textit{Medium}, I = \textit{Light} (see Figure \ref{hsimodel}).

\begin{table}[t]
\begin{tabular}{lll}
\hline
Fuzzy variable & Term set                                                          & Domain           \\ \hline
Hue            & T = \{ Red, Orange, Yellow, Green, Cyan, Blue, Violet, Magenta \} & X = {[}0, 360{]} \\
Saturation     & T = \{ Low, Medium, High \}                                       & X = {[}0, 100{]} \\
Intensity      & T = \{ Dark, Deep, Medium, Pale, Light \}                         & X = {[}0, 255{]} \\ \hline
\end{tabular}
\caption{Description of fuzzy attributes of the fuzzy color space we proposed in earlier works \cite{thesis}, \cite{jaciii}, \cite{scis}. }
\label{tab:fhsi}
\end{table}

% \hl{our previous works are here. Fuzzy color space, fuzzy color dif, fuzzy harmony, 2022 paper CA, CD. Example 1 fuzzy color and meaning, 1 fuzzy palette and meaning}

% \hl{measure of similarity between fuzzy palettes? use my existing or find better. Formulas, so we can cite later}
\section{Methods}

\begin{figure}[t]
\includegraphics[width=\textwidth]{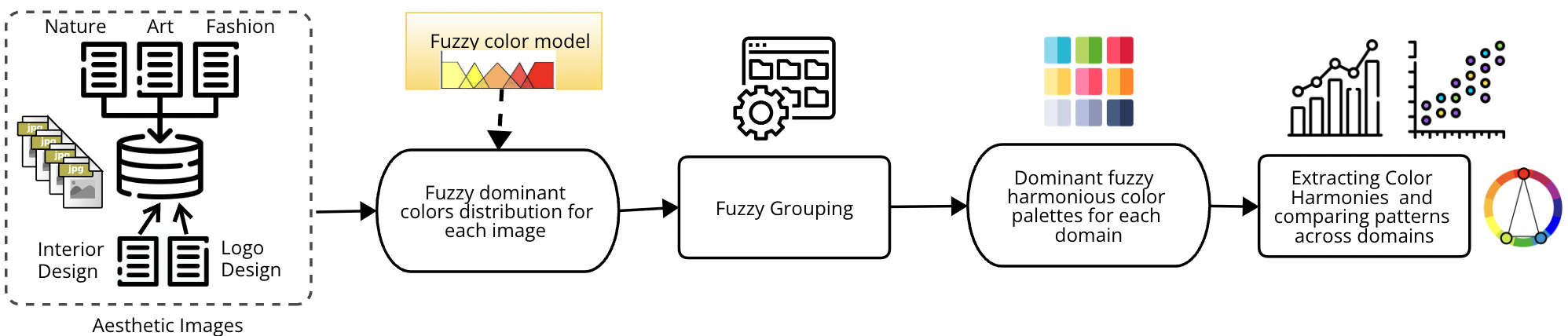}
\caption{Proposed fuzzy approach for color harmony universality estimation}
\label{fig:methodology}
\end{figure}

The proposed approach is schematically shown in Figure \ref{fig:methodology}. First, we collect a dataset comprising aesthetically appealing images from five distinct domains. Then, we extract fuzzy dominant colors in each image and group the images, forming fuzzy color palettes for each domain. Finally, we extract color harmony patterns and compare them.% across domains.

\subsection{Data Collection and Description} %Or just Data
% \hl{how collected (or the source), how many images, (may be table?}
\subsubsection{Fashion}

The dataset and fuzzy color-based palettes presented in \cite{fss} were utilized. The dataset comprises looks from a variety of sources, including prominent fashion websites like lookbook.nu, instyle.com, and dailylook.com, as well as different style communities on social networking sites (e.g., Instagram, VK, etc.).

\subsubsection{Art}
The experiment utilized a total of 1276 artworks from the 'Best Artworks of All Time' dataset \cite{kaggle_art}. This dataset comprises famous pieces of art created by various artists representing diverse movements and eras.

\subsubsection{Nature}
We used a dataset of pictures of natural landscapes \cite{nature_kaggle} to extract fuzzy palettes present in various types of landscapes. This dataset includes real-world photos from Flickr, consisting of 100 desert pictures and 184 pictures for each of the following categories: landscapes, mountains, seas, beaches, islands, and Japan. In total, we collected 1204 images.

\subsubsection{Interior Design}
We utilized a dataset of Modern Architecture consisting of 100,000 images \cite{kaggle_int}. Specifically, we focused on the Private Apartments section (U-W) and excluded images containing keywords like 'Garden,' 'Exterior,' 'Facade,' etc., as our interest was solely in interior design. This led to a total collection of 1250 interior design images.

\subsubsection{Brand Logos}
For dominant harmonious palettes in marketing, we employed the Popular Brand Logos image dataset \cite{kaggle_logo}, which comprises vector images of numerous well-known brand logos. A total of 1250 popular brand logos were utilized in the experiment.
\subsection{Color Wheel}
\begin{figure}[tb]
\includegraphics[width=0.7\textwidth]{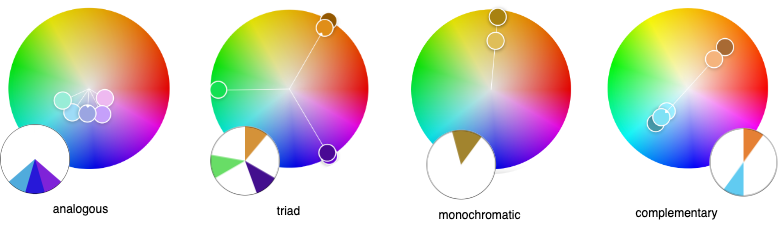}
\caption{Color harmony rules}
\label{fig:harmony}
\end{figure}

% \hl{Describe color wheel and color harmonies (triadic, analogous, etc. own figure?). "Monochromatic", "Complementary", "Split Complementary", "Triad", "Square", "Rectangular", "Analogous"}
Johannes Itten, in his book 'The Art of Color' \cite{itten}, proposed a color wheel and described several principles - graphical schemes for constructing harmonious color combinations (Figure \ref{fig:harmony}) . For instance, using a monochromatic color scheme means selecting one hue and its darker and lighter variations. Diametrically opposed colors are called complementary and produce the highest possible contrast. A split complementary color scheme involves one base color and two secondary colors. The triad scheme employs three colors that form a perfect triangle on the color wheel, while the square and rectangular harmonies follow the same logic. The analogous scheme entails selecting from 3 to 5 adjacent colors. However, it's essential to be mindful of the saturation and lightness levels to create a well-balanced and harmonious color palette. The more colors used, the more challenging it becomes to attain aesthetic pleasantness. In our experiment, we use \textit{"Monochromatic", "Complementary", "Split Complementary", "Triad", "Square", "Rectangular", "Analogous"} color relationships.

%https://drive.google.com/file/d/1C1UFMoz8528Pd6la_SHbVp8eK2EY7nEt/view?usp=sharing

% \subsection{Fuzzy Sets and Logic}
% \hl{Smth not included in Research Background. May be fuzzy rules}
%\subsection{Proposed approach}
% \hl{Formal staff - Domain D, Palette P}
% \hl{Methodology figure}

%Images were Preprocessed - resized.
\subsection{Fuzzy Palettes Extraction}

\begin{figure}[bt]
  \begin{subfigure}{0.55\textwidth}
    \includegraphics[width=\linewidth]{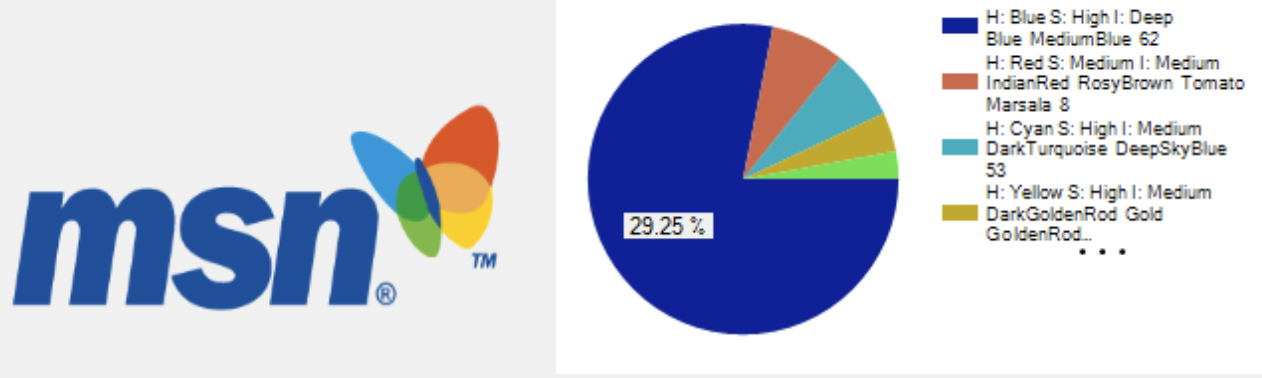}
    \caption{Logo design} 
  \end{subfigure}%
  \hspace*{\fill}   % maximize separation between the subfigures
  \begin{subfigure}{0.45\textwidth}
    \includegraphics[width=\linewidth]{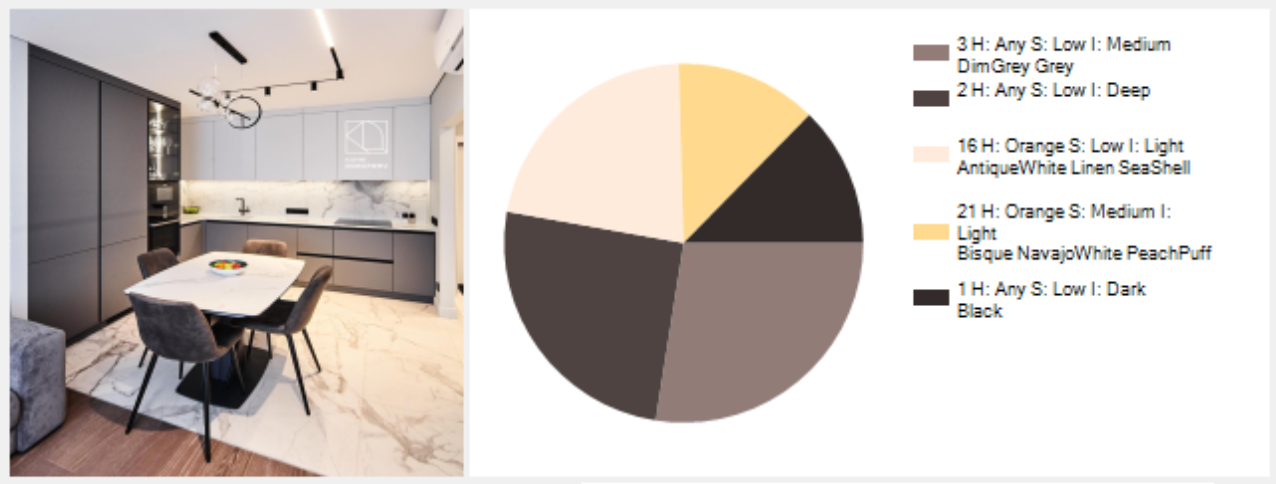}
    \caption{Interior design} 
  \end{subfigure}%
  \caption{Examples of extracted fuzzy color palettes } 
\label{fig:extraction_ex}
\end{figure}

Harmonious fuzzy color-based palettes were generated from the dataset by categorizing images with similar color compositions. Examples illustrating the fuzzy dominant colors detection from image are presented in Figure \ref{fig:extraction_ex}. We employed the fuzzy color model, along with formulas for fuzzy color difference and palette similarity (as described in \cite{jaciii}, \cite{ijufks}, and \cite{fss}). Below, we outline Algorithm \ref{alg} for identifying a set of dominant fuzzy color palettes $P_{1},...,P_{k}$ within a given domain $D$. This algorithm employs a method for assessing the similarity between two images using FHSI, denoted as $M_{1}$ and $M_{2}$ \cite{thesis}. For a more detailed explanation of the algorithm, please refer to \cite{thesis}.

We calculated the average number of images per group for each domain. Out of all the extracted palettes, we selected dominant palettes based on the criterion that they contained more images than the calculated average for each group.

\begin{algorithm}
 \KwData{dataset of images $ M_{1},...,M_{n} $ in some domain $D$}
 \KwResult{list of fuzzy dominant color palettes $ P_{1},...,P_{k} $ in $D$}
 $FuzzyPalettes$ $\leftarrow$ an empty list;
 \While{not at end of dataset}{
 read current image $ M_{i} $; \\
  $FP_{i} $ $\leftarrow$ FindFuzzyDomColors ($ M_{i} $);\\
  $Dp_{avg}$  $\leftarrow$ FindAvgPercDif ($ M_{i} $);\\
  ...
  \tcc {the perceptual difference $Dp_{avg}$ is found between $FP_{i} $  and members of each fetched harmonious group. See Algorithm 1 in \cite{fss}.}
\eIf{minimal $Dp_{avg}$ $\geq$ diffThreshold}
{form a new Palette and add $M_{i}$ to it. Add Palette to $FuzzyPalettes$}
   {
    add $M_{i}$ to a palette in $FuzzyPalettes$ with which $M_{i}$ has minimal $Dp_{avg.}$}
  }

 \Return $FuzzyPalettes$;
 \caption{Extracting fuzzy dominant palettes}
\label{alg}
\end{algorithm}

%\subsection*{Dataset Annotation on an 11-point scale}\cite{}
% \subsubsection{Fuzzy rules or CH metric}
% 1.Color wheel scheme 2.same intensity, low intensity, high intensity ? 3. low saturation ? 4. Same nuance? 5. Small number of colors? 6. Ration vs intensity Itten :Itten’s idea: Colors should be combined such that the ratio of their areas is inversely proportional to the ratio of their intensities. if std deviation of intensity is Low then similar intensity
% Ex: if color wheel correspondence then CH is HIGH
% if AVG SAT 1 is HIGH  (or SAT1 HIGH and SAT 2 HIGH)  then CH is LOW

\section{Experimental Results}
% \subsection{Results ...}
%discrepance diversity

\begin{figure}[t]
\includegraphics[width=\textwidth]{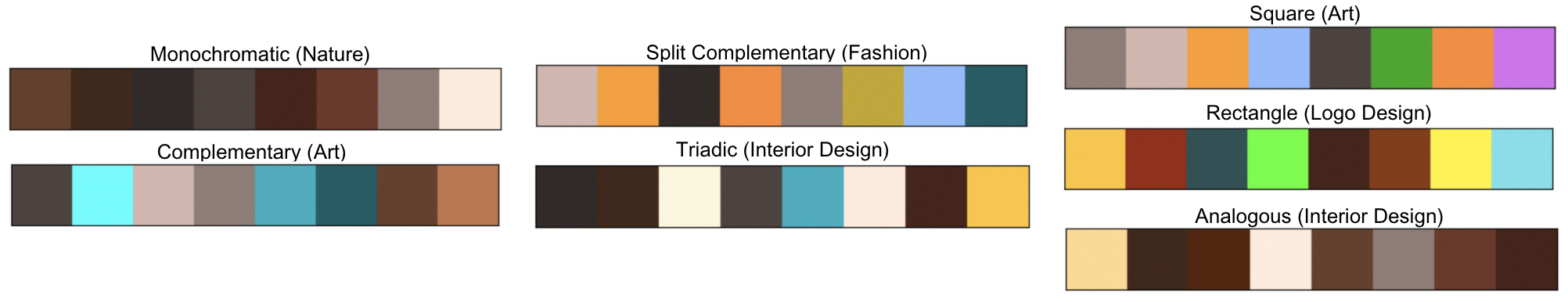}
\caption{Examples of color palettes associated with certain harmonies}
\label{fig:palettes_examples}
\end{figure}

\begin{figure}[t]
  \begin{subfigure}{0.33\textwidth}
    \includegraphics[width=\linewidth]{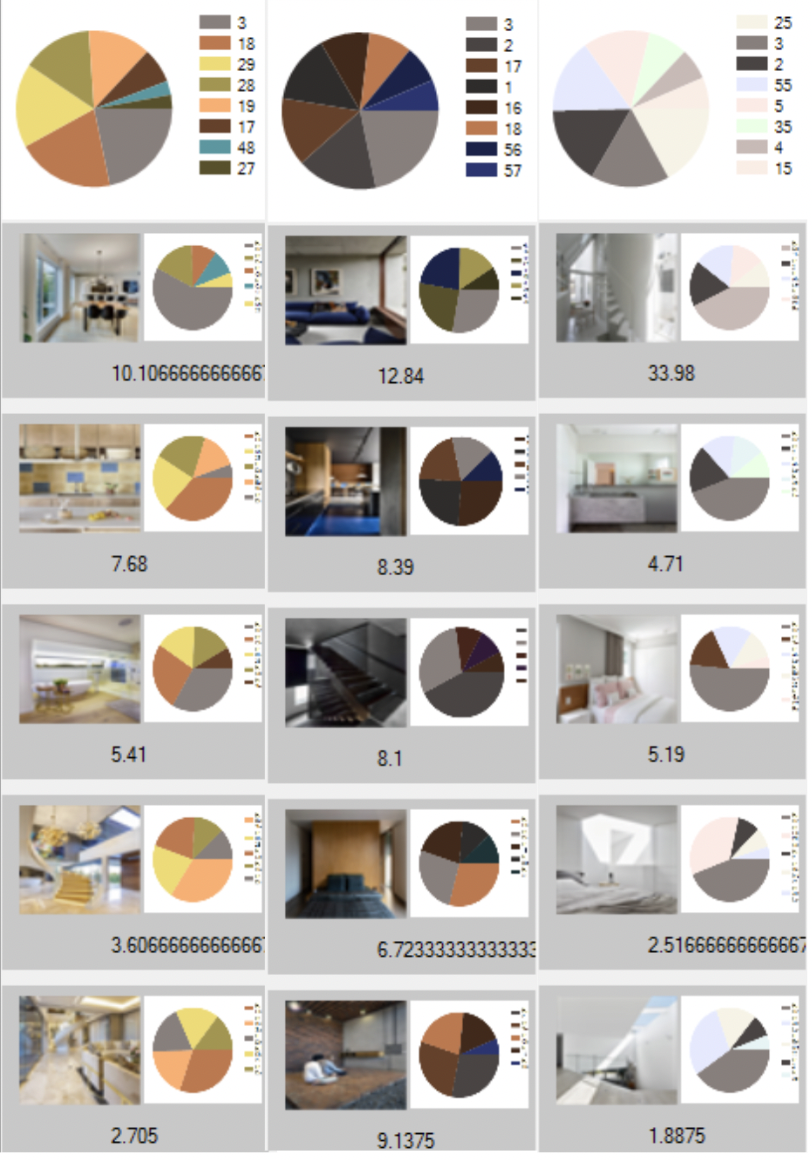}
    \caption{Interior} 
  \end{subfigure}%
  \hspace*{\fill}   % maximize separation between the subfigures
  \begin{subfigure}{0.33\textwidth}
    \includegraphics[width=\linewidth]{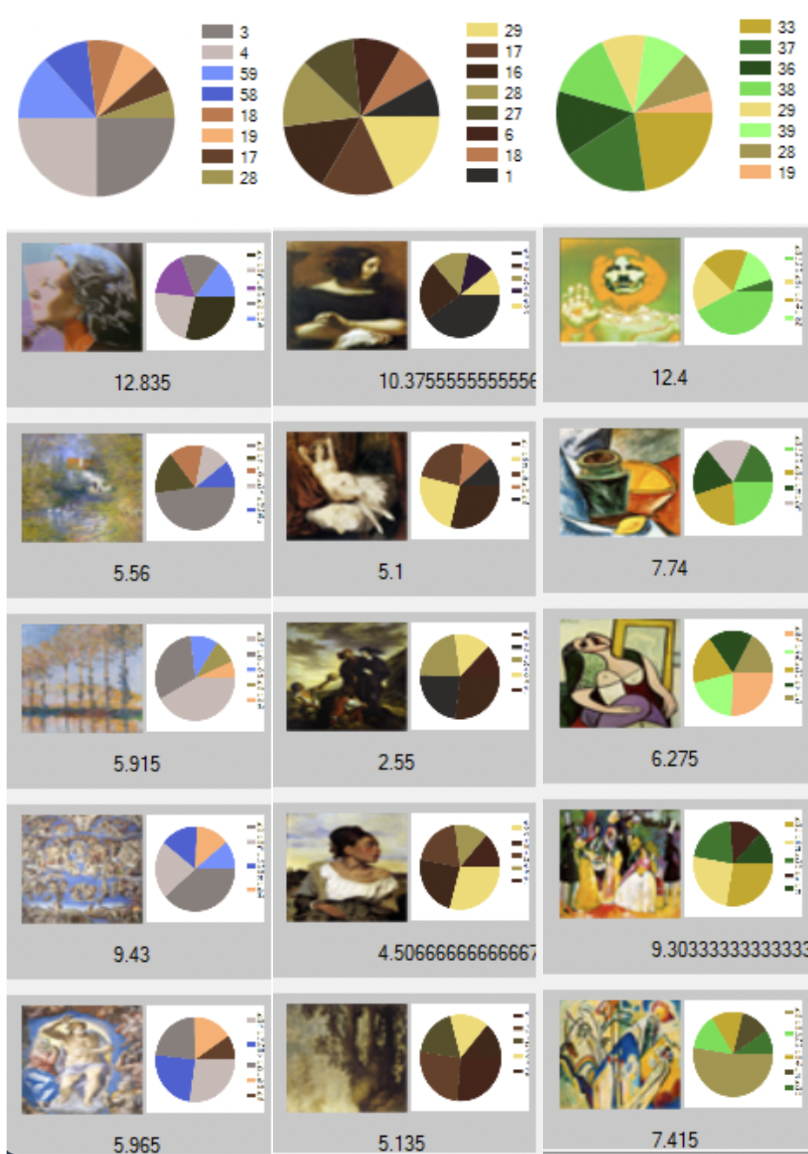}
    \caption{Art} 
  \end{subfigure}%
  \hspace*{\fill}   % maximizeseparation between the subfigures
  \begin{subfigure}{0.33\textwidth}
    \includegraphics[width=\linewidth]{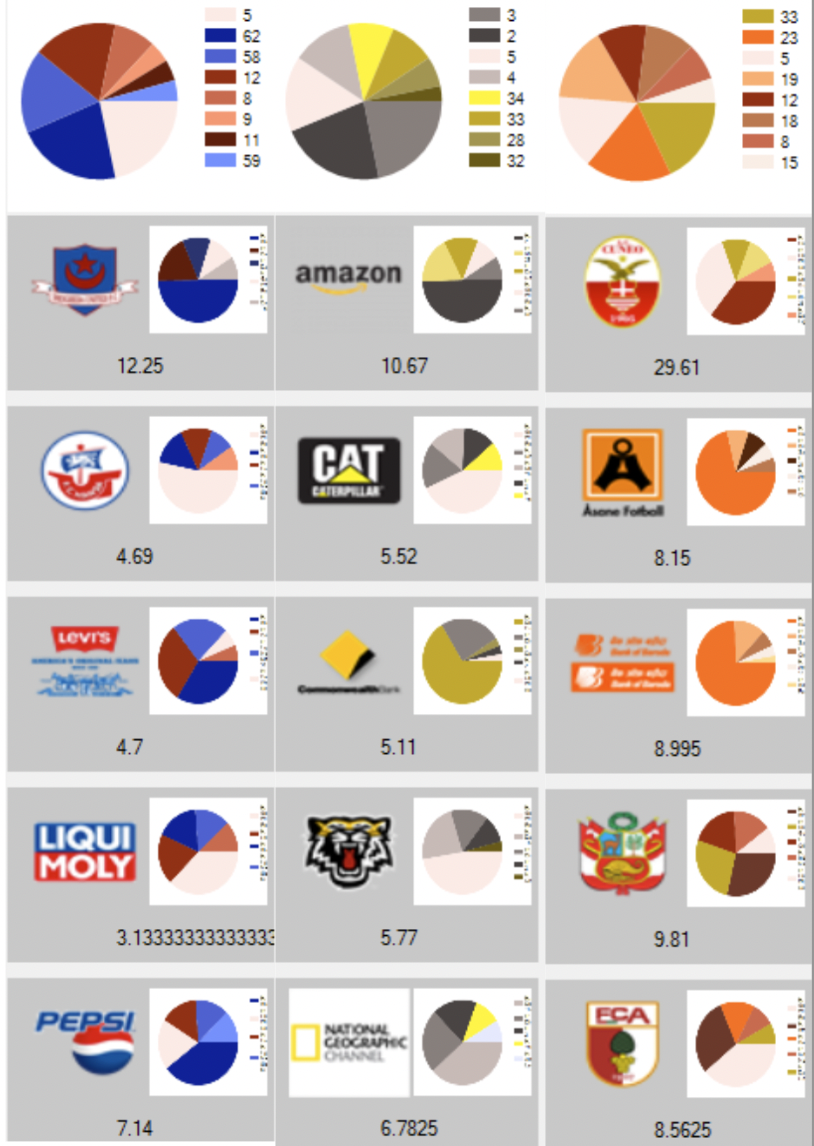}
    \caption{Logo} 
  \end{subfigure}
    \begin{subfigure}{0.66\textwidth}
    \includegraphics[width=\linewidth]{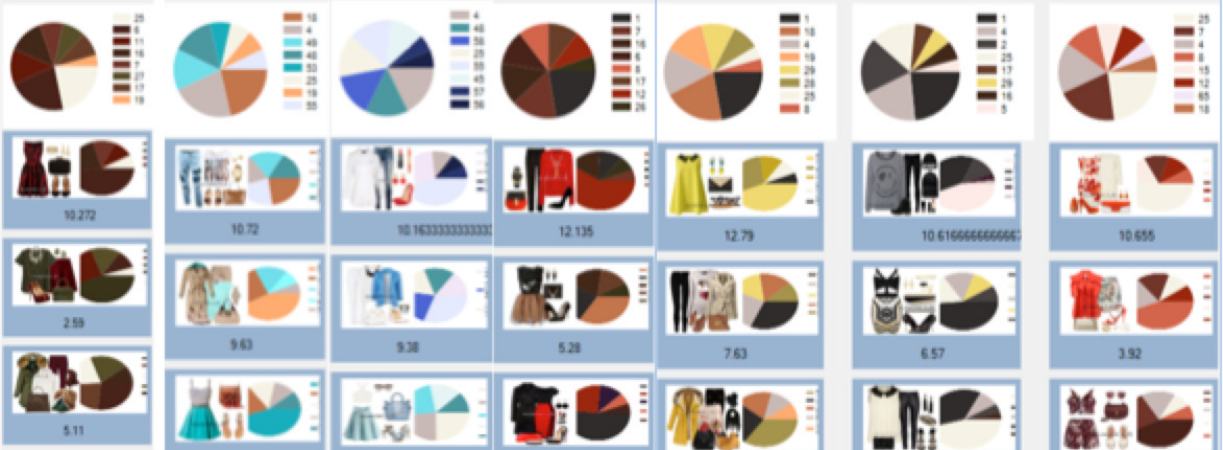}
    \caption{Fashion} 
  \end{subfigure} \hspace*{\fill} 
    \begin{subfigure}{0.32\textwidth}
    \includegraphics[width=\linewidth]{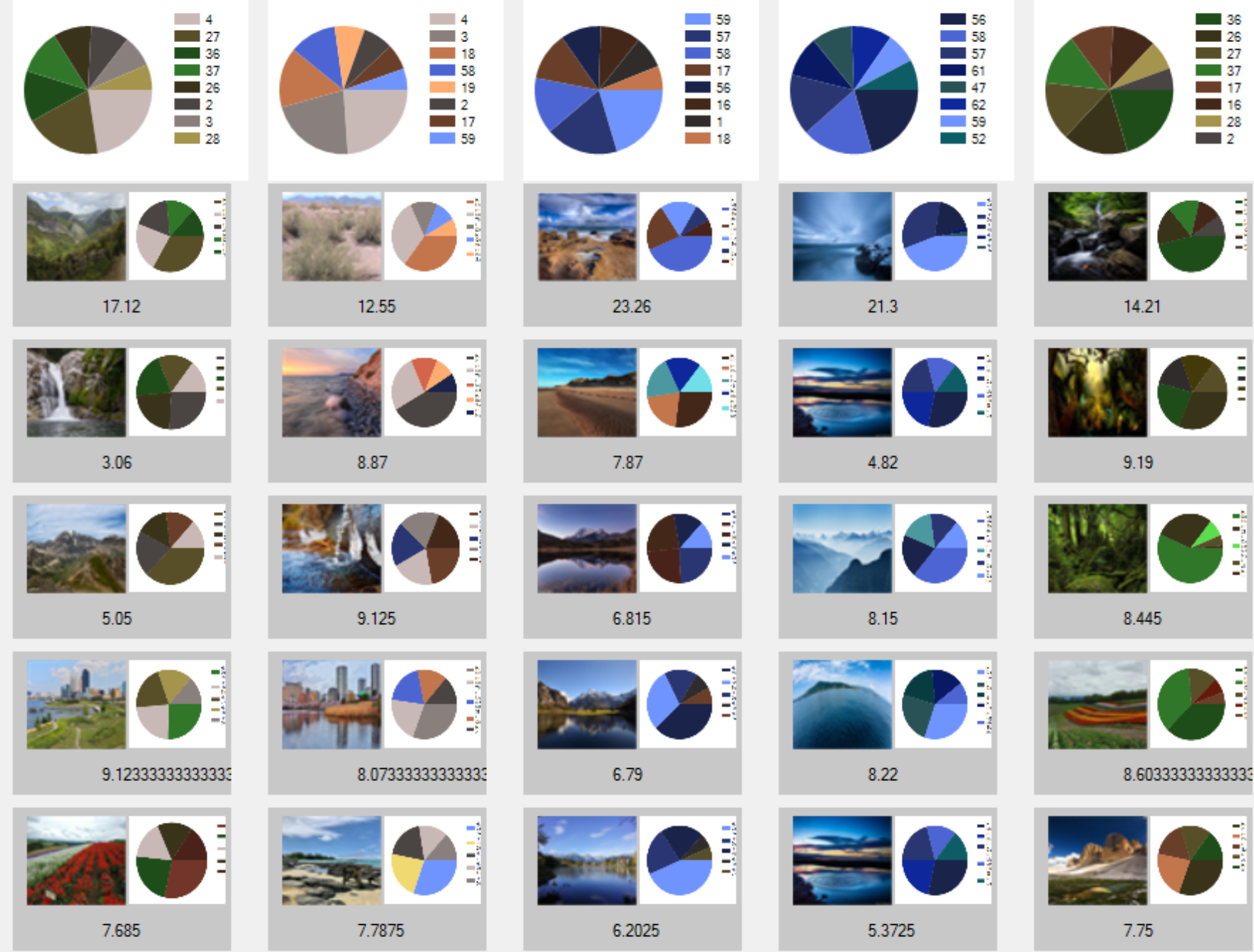}
    \caption{Nature} 
  \end{subfigure}
\caption{Examples of fuzzy dominant palettes and representative images extracted from considered domains } 
\label{many_examples_palettes}
\end{figure}

%processing
We processed datasets using Algorithm \ref{alg} to obtain fuzzy color palettes for each context. For instance, in the \textit{Art} domain, we obtained 46 groups of palettes with similar fuzzy color schemes. Figure \ref{fig:palettes_examples} showcases examples of color palettes associated with specific harmonies. Additionally, Figure \ref{many_examples_palettes} displays examples of fuzzy dominant palettes from various domains, accompanied by representative images.

\begin{table}[tb]
\begin{tabular}{lllllll}

\hline
Context & \begin{tabular}[c]{@{}l@{}}\#Dominant\\ palettes\end{tabular} & \begin{tabular}[c]{@{}l@{}}Dominant \\ harmony\end{tabular} & \begin{tabular}[c]{@{}l@{}}Unrecognized\\ Harmonies, \%\end{tabular} & \begin{tabular}[c]{@{}l@{}}Mean I\end{tabular} & \begin{tabular}[c]{@{}l@{}}Mean S\end{tabular} & \begin{tabular}[c]{@{}l@{}}Top Fuzzy Colors\end{tabular} \\ \hline
    Fashion    &  59                                                            &        Analogous                                                     &        6.8                                                              &    0.50                                              &         0.40                                         &                 \raisebox{-0.4\totalheight}{\includegraphics[width=0.15\textwidth]{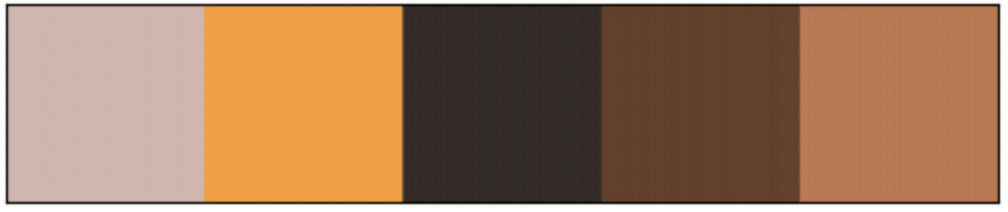}}
                                            \\
     Nature    &     62                                                          &       Complimentary                                                        &      6.5                                                                &      0.53                                             &          0.46                                        &        \raisebox{-0.4\totalheight}{\includegraphics[width=0.15\textwidth]
      {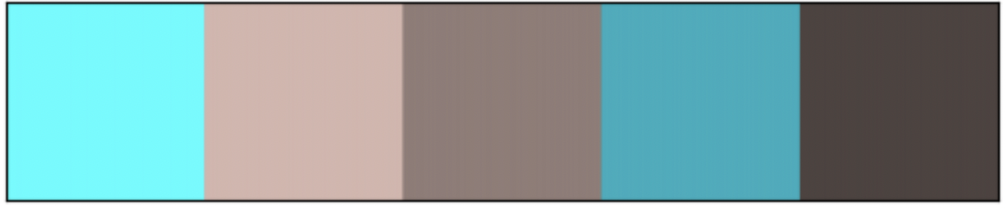}}                                                         \\
   Logo Design     &          34                                                     &       Analogous                                                        &        2.9                                                              &            0.49                                       &            0.48                                      &                \raisebox{-0.4\totalheight}{\includegraphics[width=0.15\textwidth]
      {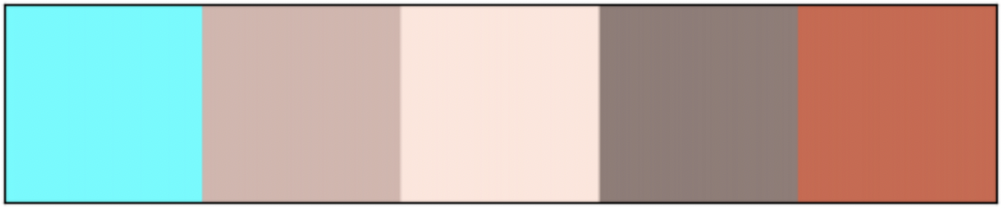}}                                                                         \\
Interior Design        &    37                                                           &         Analogous                                                      &         0                                                             & 0.47                                                  &  0.36                                                &                                      \raisebox{-0.4\totalheight}{\includegraphics[width=0.15\textwidth]
      {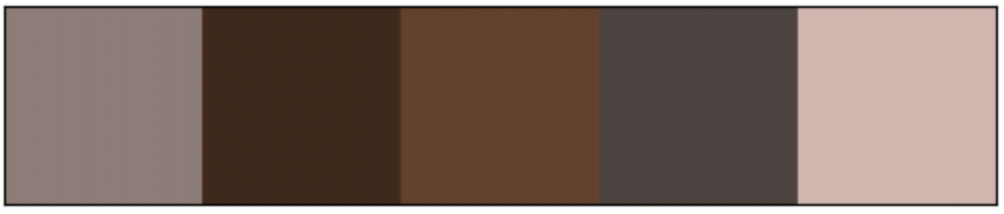}}                                             \\
      Art Images   &    46                                                           &      Analogous                                                         &        0                                                              &      0.46                                             &     0.40                                             &       \raisebox{-0.4\totalheight}{\includegraphics[width=0.15\textwidth]
      {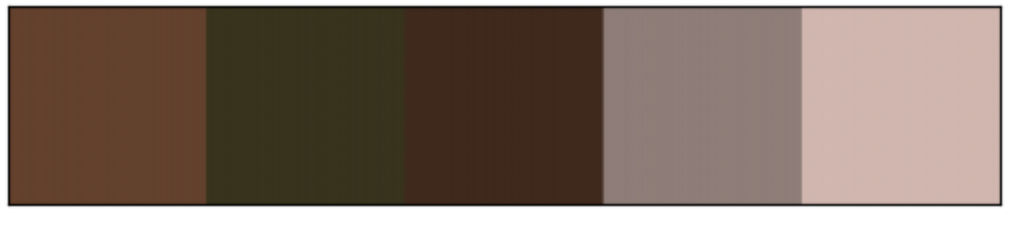}}                                                         \\ \hline
      
\end{tabular}
\caption{Summary of harmonious dominant fuzzy palettes from considered domains}
\label{tab:main}
\end{table}
%wheel
% Next, we identified colors on the RGB color wheel up to tertiary hues (12-split) and subsequently computed the associated color harmonies. The comparison results are given in Table \ref{tab:main} and shown graphically in Figures \ref{wheel_harmonies} and \ref{fig:intensities}, \ref{fig:scatter}. The majority of color schemes we extracted followed the color wheel relationship, but some did not (\textit{Other} category). Some of the rules based on a color wheel are not frequently met, like \textit{Triad, Square,} and \textit{Rectangle}. On the other hand, harmonies of \textit{Analogy} and \textit{Contrast} (Complementary) prevail in each considered domain. Note that a single palette could encompass multiple harmony schemes, such as \textit{Analogous} and \textit{Triadic}, due to our use of palettes containing eight colors. 

\begin{figure}[t]
  \begin{subfigure}{0.33\textwidth}
    \includegraphics[width=\linewidth]{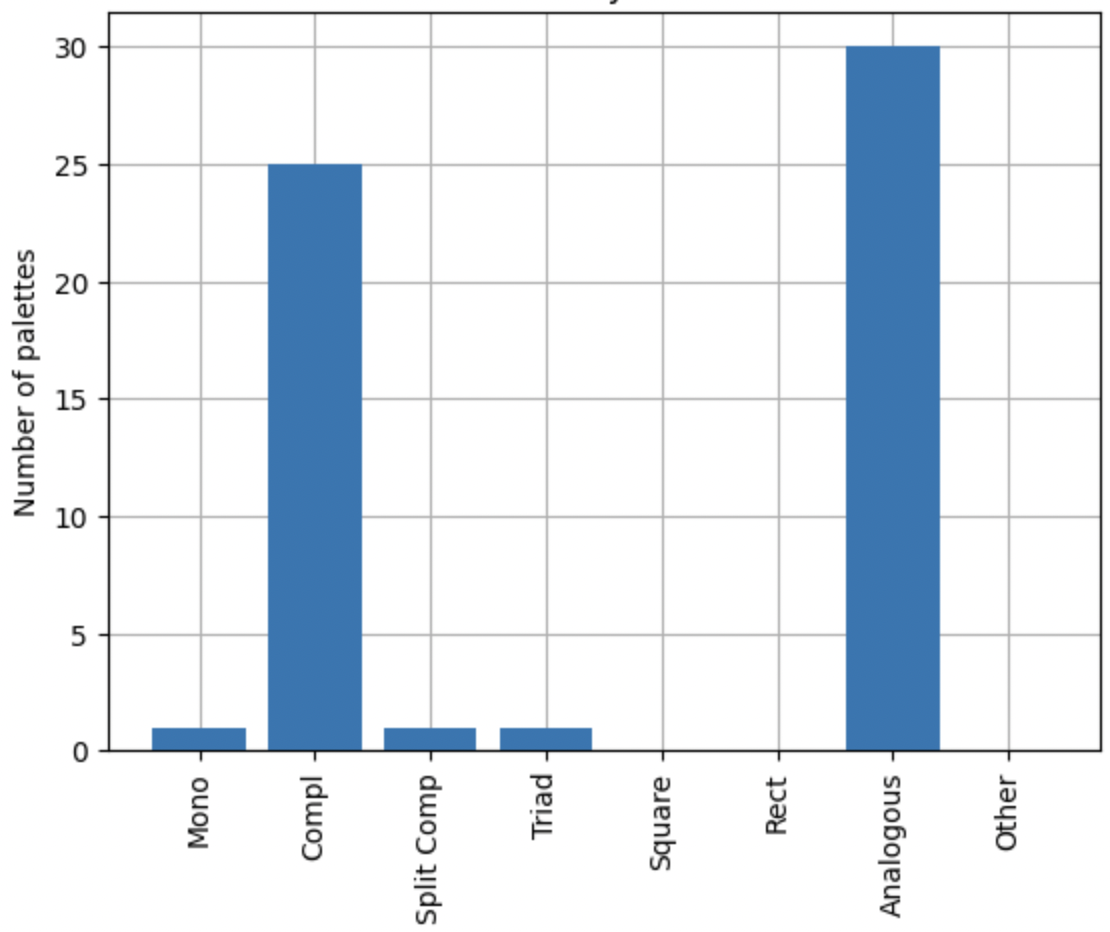}
    \caption{Interior design} 
  \end{subfigure}%
  \hspace*{\fill}   % maximize separation between the subfigures
  \begin{subfigure}{0.33\textwidth}
    \includegraphics[width=\linewidth]{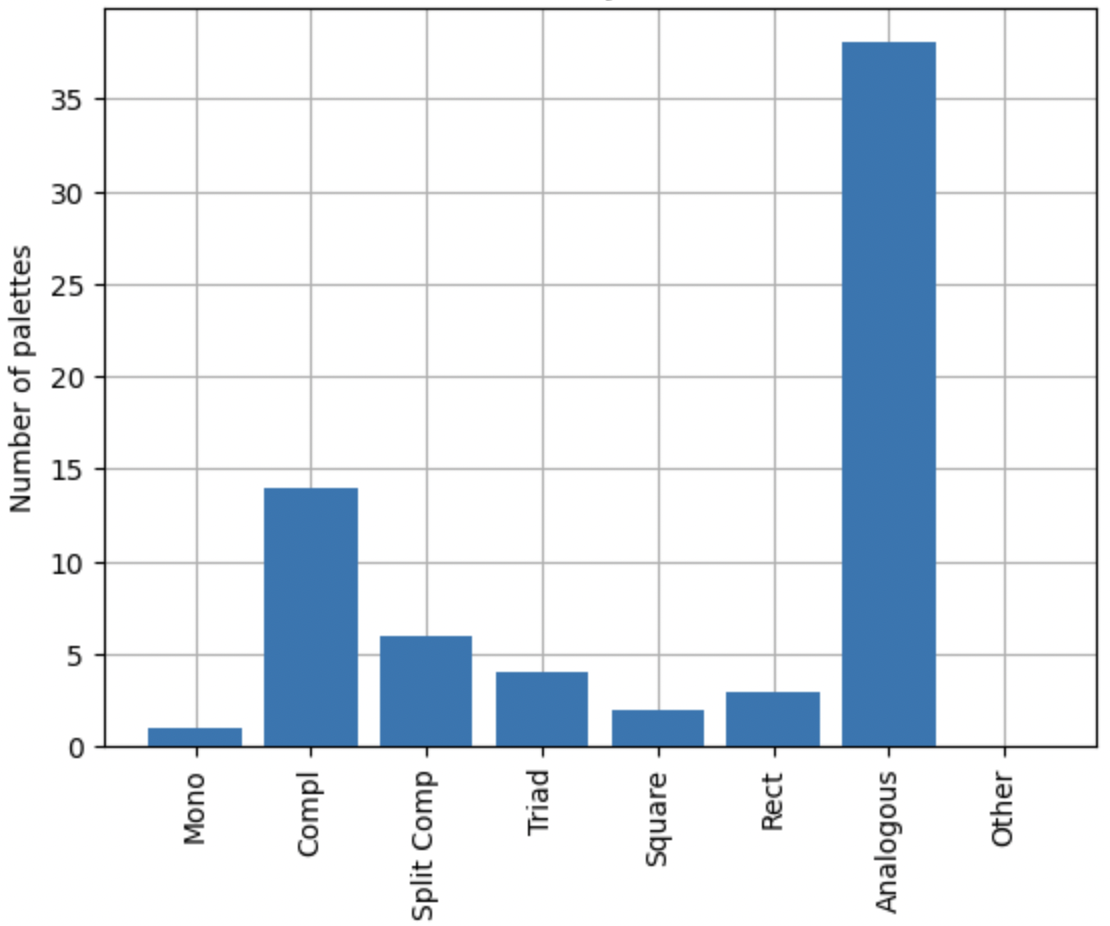}
    \caption{Art images} 
  \end{subfigure}%
  \hspace*{\fill}   % maximizeseparation between the subfigures
  \begin{subfigure}{0.33\textwidth}
    \includegraphics[width=\linewidth]{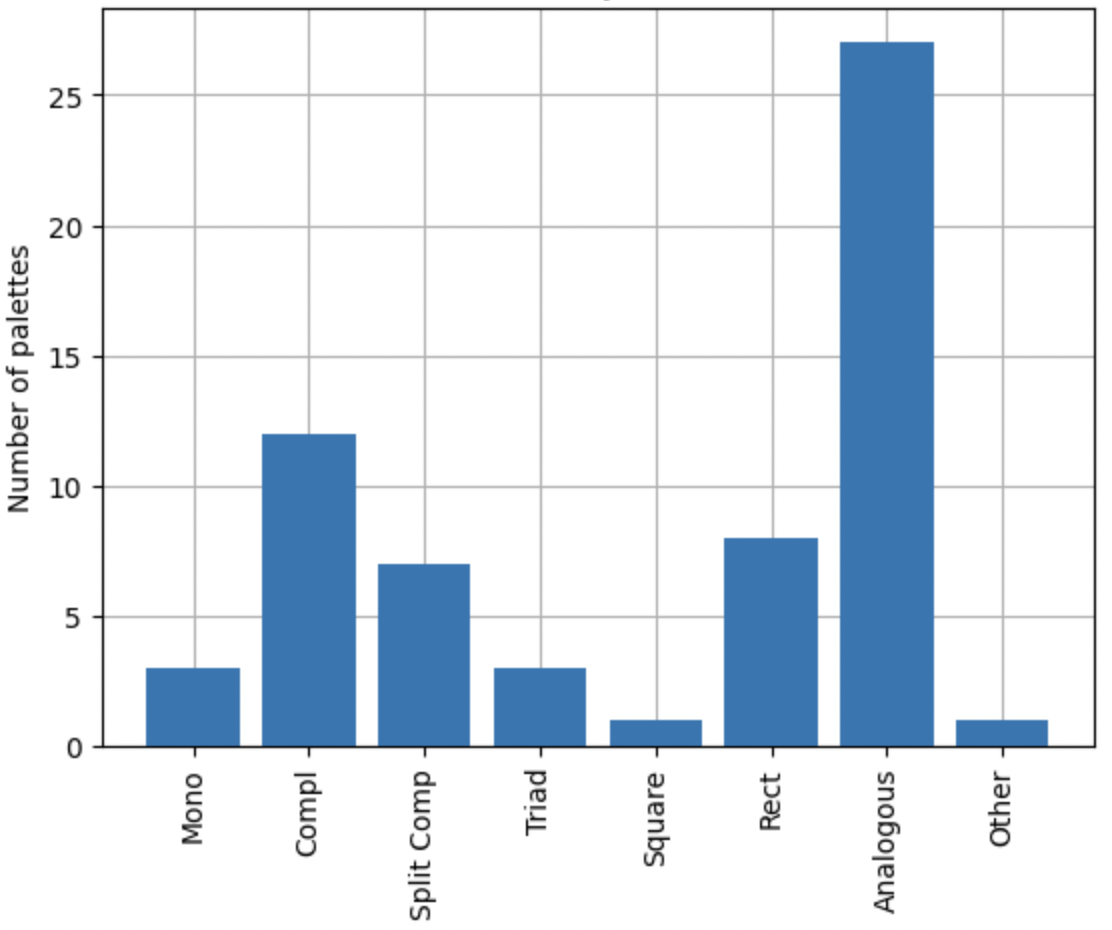}
    \caption{Logo design} 
  \end{subfigure}
    \begin{subfigure}{0.33\textwidth}
    \includegraphics[width=\linewidth]{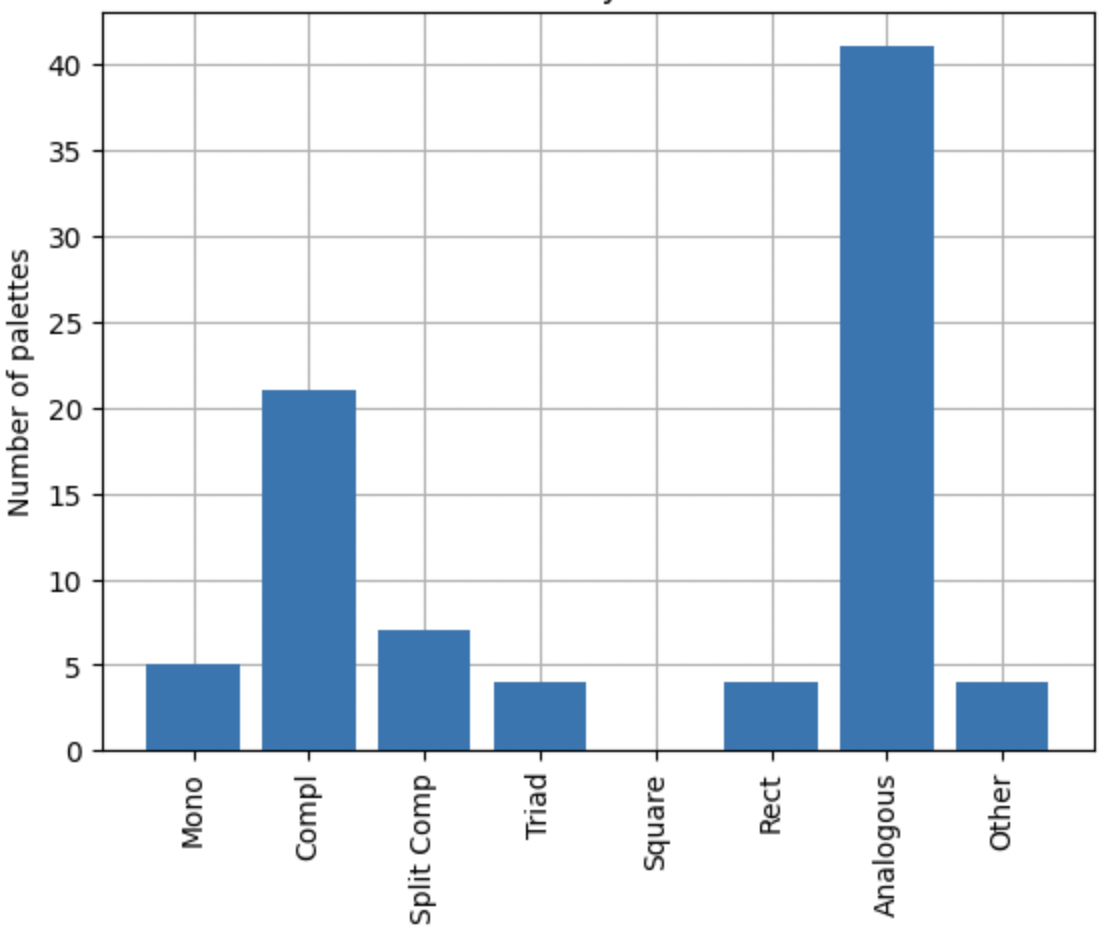}
    \caption{Fashion} 
  \end{subfigure} 
    \begin{subfigure}{0.33\textwidth}
    \includegraphics[width=\linewidth]{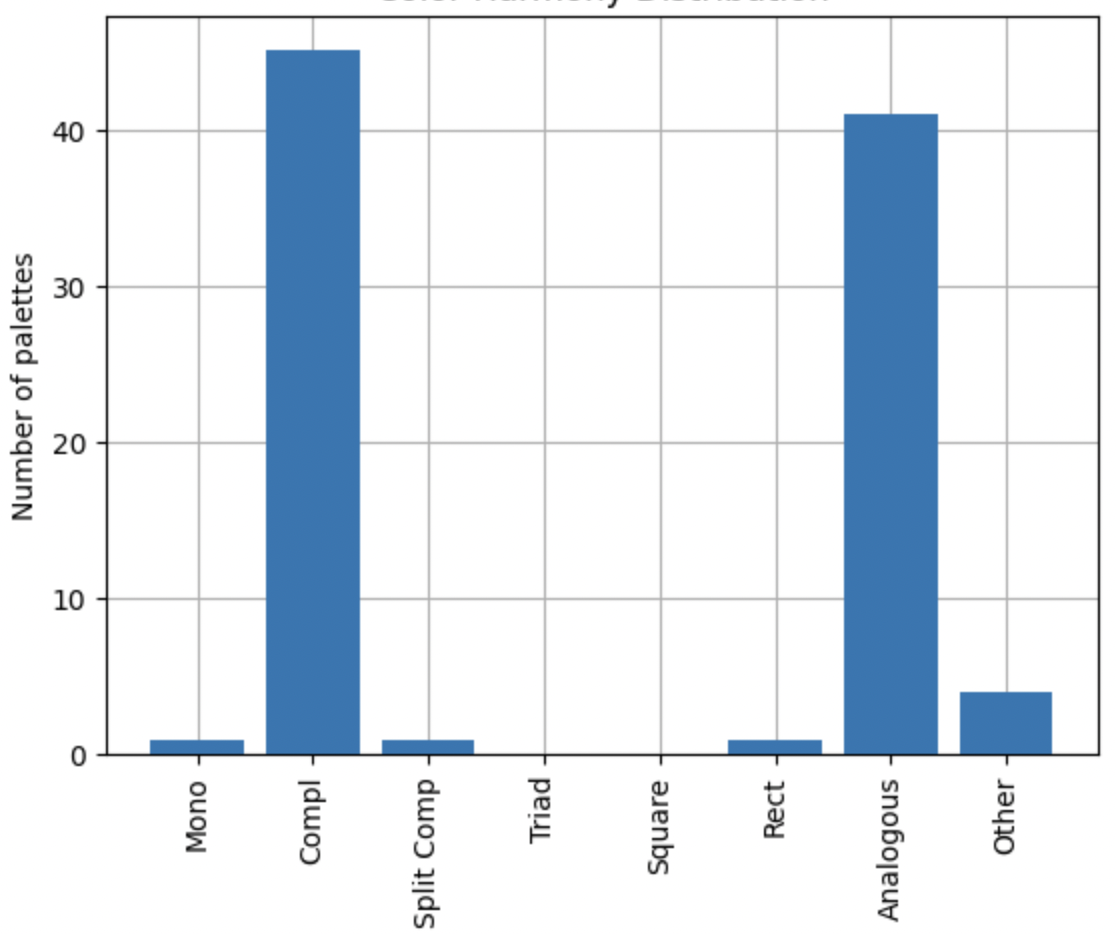}
    \caption{Nature} 
  \end{subfigure}
     
\caption{Distribution of Color Harmonies among considered domains } 
\label{wheel_harmonies}
\end{figure}

\begin{figure}[tb]
  \begin{subfigure}{0.57\textwidth}
    \includegraphics[width=\linewidth]{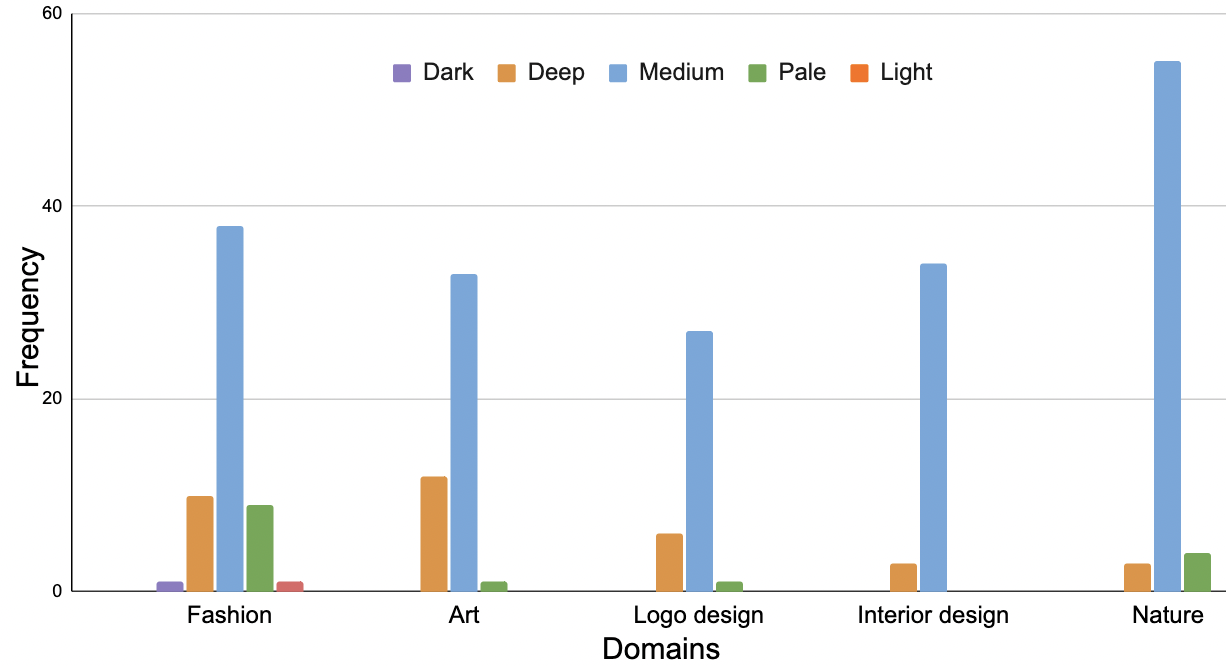}
    \caption{Distibution of fuzzy intensities across domains}
    \label{fig:intensities}
  \end{subfigure}%
  \hspace*{\fill}   % maximize separation between the subfigures
  \begin{subfigure}{0.43\textwidth}
    \includegraphics[width=\linewidth]{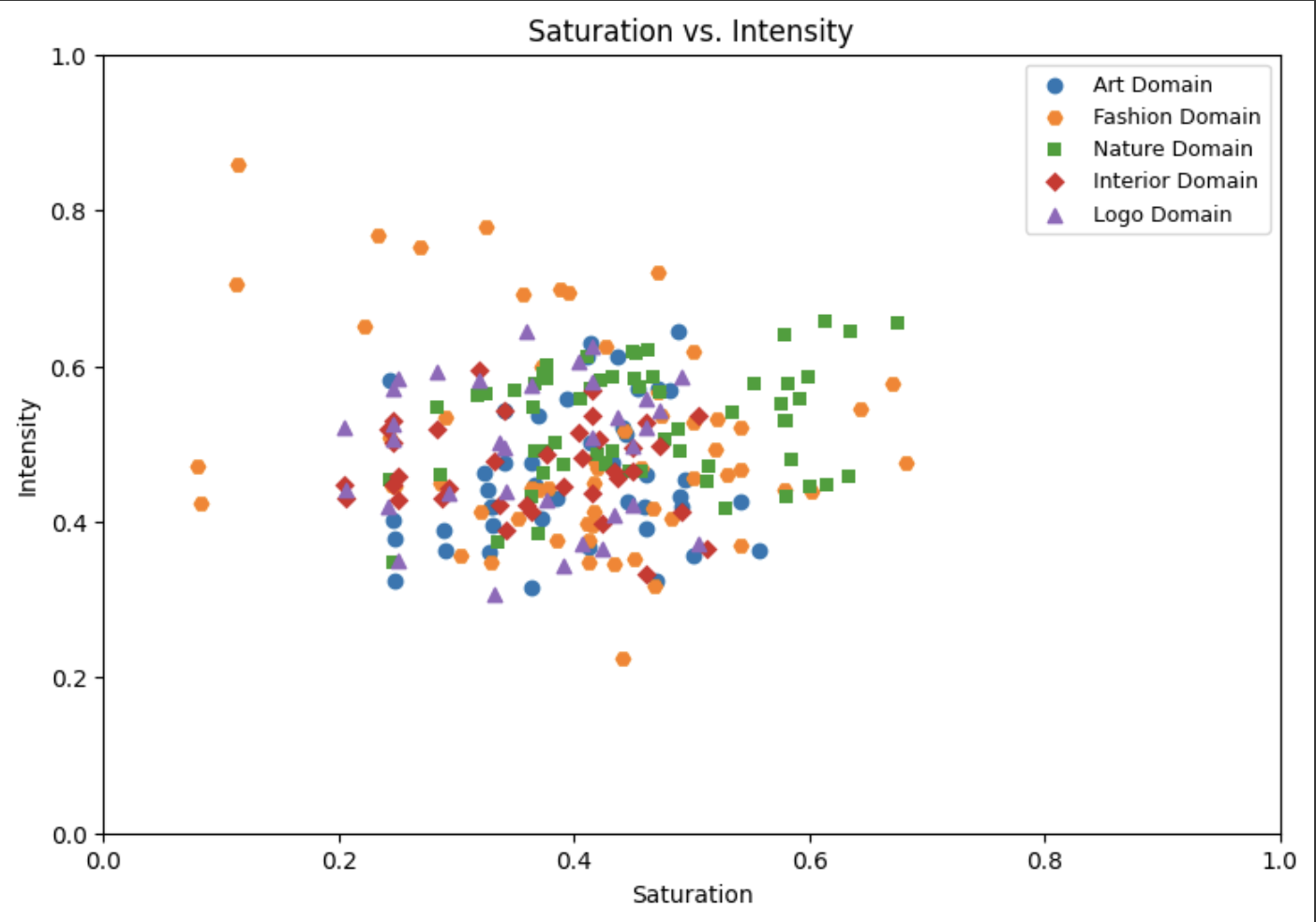}
    \caption{Trends in \textit{Intensity} and \textit{Saturation}} 
    \label{fig:scatter}
  \end{subfigure}%
  \caption{Distribution of intensities. Fuzzy partition \textit{(Dark, Deep, Medium, Pale, Light) } is shown in Figure \ref{hsimodel}.} 

\end{figure}

 We proceeded to identify colors on the RGB color wheel, encompassing tertiary hues (12-split), and subsequently computed the corresponding harmonies. The comparative results are detailed in Table \ref{tab:main} and visually presented in Figures \ref{wheel_harmonies}, \ref{fig:intensities}, and \ref{fig:scatter}. The majority of the extracted color schemes adhered to color wheel relationships, but some fell into the \textit{'Other'} category, deviating from these norms. Certain wheel-based rules, such as \textit{'Triad,' 'Square,'} and \textit{'Rectangle,'} were less frequently observed. In contrast, \textit{'Analogous'} and \textit{'Complementary'} harmonies predominated in each of the domains under consideration. Notably, a single palette could encompass multiple harmony schemes, such as \textit{'Analogous'} and \textit{'Triadic'}, owing to our use of palettes consisting of eight colors.

%I, S
% As can be clearly seen harmony rules based on the color wheel operate on certain \textit{Intensity} and \textit{Saturation} levels (see Figures \ref{fig:scatter} and \ref{fig:intensities}. Even when hues adhere to specific color wheel relationships, such as \textit{Triadic}, variations in saturation and intensity are observed to play a substantial impact in determining harmony levels. The majority of color schemes with strong color harmony followed color wheel rules and kept certain patterns - \textit{medium} levels of \textit{Saturation} and \textit{Intensity}. This was observed in all contexts. Overall, experimental results suggest that color harmony is largely universal but still somewhat context-dependent.

It's evident that harmony rules based on the color wheel operate on specific levels of \textit{Intensity} and \textit{Saturation}, as depicted in Figures \ref{fig:scatter} and \ref{fig:intensities}. Even when hues conform to particular color wheel relationships, such as the \textit{'Triadic'} scheme, variations in saturation and intensity significantly impact harmony levels. In the majority of color schemes exhibiting strong color harmony, adherence to color wheel rules coincided with consistent patterns of \textit{medium} levels of \textit{Saturation} and \textit{Intensity}. This pattern was observed across all contexts. Overall, the experimental results suggest that while color harmony exhibits a significant degree of universality, it remains somewhat context-dependent.

% \begin{figure}[htbp]
% \includegraphics[width=\textwidth]{fig2/freqI.png}
% \caption{Distribution of fuzzy intensities \textit{(Dark, Deep, Medium, Pale, Light) }across domains. The fuzzy partition presented in Figure \ref{hsimodel}.}
% \label{fig:intensities}
% \end{figure}

% \hl{calculate and describe correlations between each other (harmonies form different contexts). Compare datasets and their palettes : universal rules or no, which context are similar, top colors in each, top palettes, if not color wheel then what.Comparison table, dominant colors (how many in each context, average.), how manu color hues in average in each context (basic colors, there are 9-10 of them). Most frequently met couples and triples of colors. Differenc between man-made , nature colors. What is common to majority of palettes?}
% \hl{find correlations. comparisons with https://colorhunt.co/}
% \hl{comparison with RGB color harmony?pypi}

%\subsection{Accuracy Evaluation}

% \subsection{Harmony Evaluation Example}
% \hl{Example - some image - here is  palette and rules and visualization }
%\section{Discussion}

\section{Conclusion}
% \hl{Main message and main results and outcomes}
% Discussion %color wheel, I, S consider
%We can clearly see from the results that harmonious palettes followed certain intensity and saturation patterns 
The current paper explores the context dependency of color harmony using a fuzzy-based approach. We examined color harmonies across contexts of art, fashion, nature, interior design, and branding, finding that while color wheel principles play a significant role in achieving harmony, the interplay of saturation and intensity is equally vital. The majority of color schemes with strong harmony adhered to \textit{Analogous} and \textit{Complementary} color wheel rules and maintained a balance of medium saturation and intensity. 

%Our findings show that color harmony is largely universal 

Our findings underscore the largely universal nature of color harmony but also highlight its sensitivity to context. These results confirm those reported in \cite{Ou2009} regarding general patterns of color harmony. %Other findings indicate that the feeling of beauty is partially universal and domain-specific \cite{domain}. 
The study has certain limitations. The datasets used may not fully represent the diversity of real-world images and contexts. Expanding the dataset variety and size could enhance the generalizability of our findings. As for future work, we plan to introduce a fuzzy inference system for measuring color harmony. The fuzzy rules will be used to account for the color wheel correspondence and levels of saturation and intensity. We also plan to incorporate subjective user evaluations to get richer insights into color harmony. %Our work contributes to a deeper understanding of color aesthetics, offering valuable insights for designers, artists, and researchers across multiple disciplines.

%\hl{Limitations - not considering ind difference}
%The current paper examines the context dependency of essential variables affecting aesthetic pleasantness - color harmony. Our study sheds light on the nuanced interplay between hue, saturation, and intensity, 
 %Fuzzy logic allows us to model and deal with uncertainty and imprecision in human preferences and perceptions, which are often not easily quantifiable with traditional crisp logic. 

\bibliographystyle{vancouver}
\bibliography{library}

\end{document}